\pdfoutput=1

\documentclass[11pt]{article}

\usepackage{ACL2023}

\usepackage{times}
\usepackage{latexsym}

\usepackage[T1]{fontenc}

\usepackage[utf8]{inputenc}

\usepackage{microtype}

\usepackage{inconsolata}

\usepackage{graphicx}
\usepackage{amsmath}
\usepackage{booktabs}
\usepackage{tabularx}
\usepackage{multicol}
\usepackage{subcaption}
\usepackage{multirow}
\usepackage{CJK}
\usepackage{float}

\usepackage{longtable}

\title{DMoERM: Recipes of Mixture-of-Experts for Effective Reward Modeling}

\author{Shanghaoran Quan
   \\
   \\
  Beihang University \\
  \texttt{shrquan@buaa.edu.cn} \\}

\begin{document}

\maketitle
\begin{abstract}
The performance of the reward model (RM) is a critical factor in improving the effectiveness of the large language model (LLM) during alignment fine-tuning. There remain two challenges in RM training: 1) training the same RM using various categories of data may cause its generalization performance to suffer from multi-task disturbance, and 2) the human annotation consistency rate is generally only $60\%$ to $75\%$, causing training data to contain a lot of noise. To tackle these two challenges, we introduced the idea of Mixture-of-Experts (MoE) into the field of RM for the first time. We propose the \underline{D}ouble-Layer \underline{MoE} \underline {RM} (DMoERM). The outer layer MoE is a sparse model. After classifying an input into task categories, we route it to the corresponding inner layer task-specific model. The inner layer MoE is a dense model. We decompose the specific task into multiple capability dimensions and individually fine-tune a LoRA expert on each one. Their outputs are then synthesized by an MLP to compute the final rewards. To minimize costs, we call a public LLM API to obtain the capability preference labels. The validation on manually labeled datasets confirms that our model attains superior consistency with human preference and outstrips advanced generative approaches. Meanwhile, through BoN sampling and RL experiments, we demonstrate that our model outperforms state-of-the-art ensemble methods of RM and mitigates the \textit{overoptimization} problem. Our code and dataset are available at: \url{https://github.com/quanshr/DMoERM-v1}.

\end{abstract}

\section{Introduction}

After an initial stage of pre-training and subsequent instruction fine-tuning, large language models (LLMs) undergo a crucial stage of high-quality alignment fine-tuning based on Reinforcement Learning with Human Feedback (RLHF) to improve their abilities~\citep{2022Training, stiennon2020learning}. 
During the RLHF process, a reward model (RM) often needs to be trained, which acts as a proxy of human preferences and assigns scores to the outputs of the LLM. The scores are then used as reward signals to optimize the LLM through reinforcement learning (RL). In this process, the LLM and RM are interdependent and iteratively optimized, and the RM is expected to be highly consistent with human preferences. 
In addition, during the inference stage, the RM can also be augmented with Best-of-$n$ (BoN) sampling strategies to further enhance the quality of the outputs of the LLM~\citep{2022Training, 2021WebGPT}.

Training of reward models relies on the data derived from human annotators who manually rank the varying outputs under a single input by their preferences. However, many studies have found that agreement rates among human annotators typically only range between $60$-$75\%$~\citep{2019Fine, stiennon2020learning, dubois2023alpacafarm}, thereby introducing a minimum of $25\%$ noise within the labeled dataset. 
One important reason for this phenomenon is the multifaceted nature of evaluation\textemdash it is often observed that one response may excel in one aspect while simultaneously falling short in another. This multifaceted evaluation conundrum has been exemplified in previous studies~\citep{dai2023safe, ganguli2022red, bai2022training}, which illustrate the inherent tensions between enhancing helpfulness and harmlessness. As these attributes can at times be inversely related, adjudicating between a response that is more helpful yet potentially less harmless poses a significant challenge for comparative assessment. We further validate this perspective through empirical studies.



\begin{figure}[h]
  \centering
  \includegraphics[width=\linewidth]{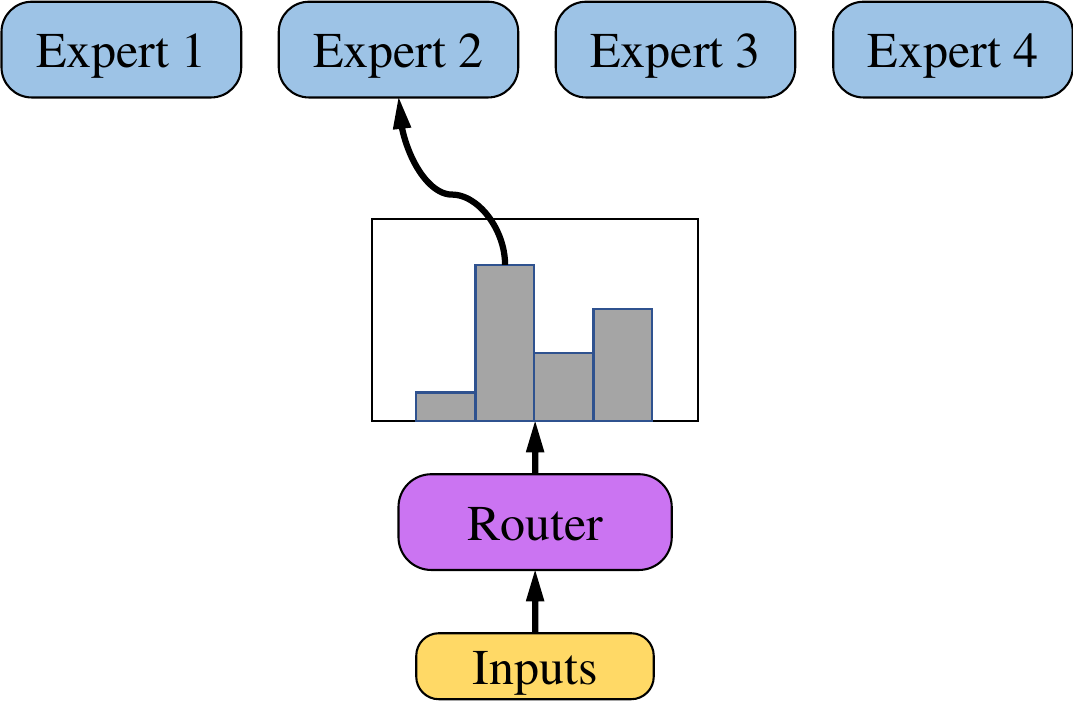}
  \caption{The outer MoE routes inputs to corresponding task-specific inner MoE.}
  \label{router}
\end{figure}


In this study, we pioneer the integration of the Mixture-of-Experts (MoE) framework~\citep{jacobs1991adaptive, lepikhin2020gshard, du2022glam} into Reward Modeling. Our approach employs a double-layer MoE architecture. The outer layer comprises a sparse MoE model specifically designed to avoid multi-task disturbance~\citep{standley2020tasks}. As shown in Figure~\ref{router}, We categorize inputs into several distinct tasks and use a pre-trained router to route the inputs to their corresponding task-specific expert. This strategy can also facilitate distributed deployment and enhance the model's capacity and capabilities without a commensurate increase in computational demands~\citep{rajbhandari2022deepspeed, xue2022go}.

Subsequently, within each inner layer lies a dense MoE model, which is tailored to the specific capabilities set required for its category. For instance, in roleplay scenarios, we divide it into six core capabilities, including personality and emotional investment, conversational sense, empathy ability and so on (details are in Appendix Table~\ref{capabilitypartition}). We obtain preference labels on these single capability points by calling a public API, which greatly reduces annotation costs and is sufficient to achieve satisfactory results in our experiments. Considering that capability points are equivalent to a decomposition of tasks in a low dimensional space, using low-rank adaptation (LoRA)~\citep{hu2021lora} fine-tuning will be very suitable. 
Each LoRA fine-tuned model effectively becomes an \textit{expert} in scoring on a singular capability point. Lastly, we aggregate the outputs from these expert models into a unified one-dimensional score with an MLP to determine the final reward value. We believe this methodology can improve the interpretability and performance of RMs since it is just like the \textit{CoT} for RMs. Both preference consistency and optimization evaluations indicate our model is more effective in optimizing LLMs and can mitigate the \textit{overoptimization} problem against other state-of-the-art RM ensemble methods.

\section{Related Work}

\subsection{Mixed-of-Experts}

Mixture-of-Experts or MoEs was introduced early for machine learning applications~\citep{jacobs1991adaptive, jordan1994hierarchical}, where researchers control the allocation of different weights to different models through gate networks to mitigate interference between different types of samples. MoE can enhance the model's generalization capability by decomposing complex tasks into several subtasks, which will help avoid multi-task disturbance~\citep{standley2020tasks} and meanwhile confer greater flexibility in development~\citep{DBLP:conf/kdd/MaZYCHC18}. Recently, much ongoing research has focused on the top-$k$ (e.g., top-$1$ or top-$2$ in many works) activation MoE model~\citep{ramachandran2018diversity, clark2022unified, dai2022stablemoe}, since it can be leveraged to enlarge parameter count and enhance model capability while keeping computational complexity nearly unchanged for both training and inference due to its sparse activating nature~\citep{shazeer2017outrageously, fedus2022switch}. While MoE has achieved great success in the field of large generative language models~\citep{shen2023mixture, achiam2023gpt}, how to efficiently train more effective RM with MoE architecture remains largely unexplored.


\subsection{Reward Model Ensembling}

Reward Model Ensembling has been tried in the field of safe RLHF~\citep{dai2023safe}. The research is based on a widely observed imagination: the pursuit of greater helpfulness and harmlessness may often conflict in practice~\citep{ganguli2022red, bai2022training}. Another weave of research ensemble RMs through multi-objective reward modeling~\citep{DBLP:conf/nips/RameCDGSSC23, DBLP:conf/icml/RameAZCBL23} or weight-averaged reward modeling~\citep{DBLP:journals/corr/abs-2401-12187}, but they are struggling to formulate non-linear relationships. In addition, some research works~\citep{coste2023reward, eisenstein2023helping, zhai2023uncertainty} have found that training multiple reward models and aggregating them by changing the data training order, batch size, and learning rate can alleviate the overoptimization problem~\citep{gao2023scaling} of RM and increase its performance. However, the aggregation methods they chose were only 1) mean, 2) min, and 3) mean minus std, and the performance of aggregation was very dependent on the diversity of several models~\citep{zhai2023uncertainty}, which required a lot of attempts in experiments. 

\section{Empirical Study}

\subsection{Multi-Task Training}
\label{sec:DiffCat}
It has been frequently observed that using irrelevant training data to train LLMs will cause their generalization performance to decrease in other tasks~\citep{dong2023abilities, wen2023chathome}. This also applies to RMs. \citet{dai2023safe} found that training RMs on harmlessness and helpfulness simultaneously will lead to the model getting suboptimal results on both types of data. In order to further explore whether data of different categories interfere with each other, we selected the preference data for three tasks: roleplay, chitchat, and text creation. We train on different combinations of training sets and testing on all test sets. The results are shown in Table~\ref{differentset}.

\begin{table}[htbp]
  \centering
  \begin{tabular}{c|ccc}\toprule 
  Dataset & \#A & \#B & \#C \\\midrule 
   \#A & $\mathbf{56.7\%}$ & $51.1\%$ & $49.8\%$ \\
   \#B & $52.4\%$ & $\mathbf{54.3\%}$ & $50.3\%$ \\
   \#C & $50.8\%$ & $51.0\%$ & $\mathbf{58.1\%}$ \\\midrule 
   \#A $\cup$ \#B & $54.2\%$ & $52.6\%$ & $50.2\%$ \\
   \#A $\cup$ \#C & $53.3\%$ & $50.9\%$ & $55.6\%$ \\
   \#B $\cup$ \#C & $51.7\%$ & $52.5\%$ & $56.4\%$ \\\midrule 
   \#A $\cup$ \#B $\cup$ \#C & $52.0\%$ & $51.5\%$ & $54.8\%$ \\\bottomrule
  \end{tabular}
  \caption{\label{differentset} The results of training on different combinations of training sets and testing on all test sets. \#A, \#B, and \#C represent roleplay, chitchat and text creation respectively. The best values are written in bold.}
\end{table}

We find that using separate category data give the best results in this category, and using data from other categories may affect the generalization ability under original task.

\subsection{Annotation Consistency}
The consistency rate of manually labeled preference data is generally only $60$-$75\%$~\citep{2019Fine, stiennon2020learning, dubois2023alpacafarm}, which brings a lot of noise to the training data. Inspired by the fact that Chain-of-Thought (CoT) can improve the accuracy of reasoning~\citep{wei2022chain, zhou2022least}, we try to improve the consistency rate through CoT. We first conduct experiments on humans. We select the text creation subtask and randomly divide $200$ pairs of responses into two groups. For the first group, we let three annotators directly rank the preference and record the average agreement rate between any two annotators, which is represented as A. For the second group, we divide the text creation into five capability points: 1) intent conformity, 2) expressiveness, 3) readability, 4) content richness, 5)logic, and ask the annotators to first score the content in each single capability point, and then evaluate the overall preference. The average agreement rates on five capability points are represented as B-1 to B-5, respectively, and the final overall agreement rate is B-f. We record these results in Figure~\ref{empirical2}, and a screenshot of the annotation interface is shown in Figure~\ref{screenshot} in the Appendix.

\begin{figure}[hbtp]
  \centering
  \includegraphics[width=\linewidth]{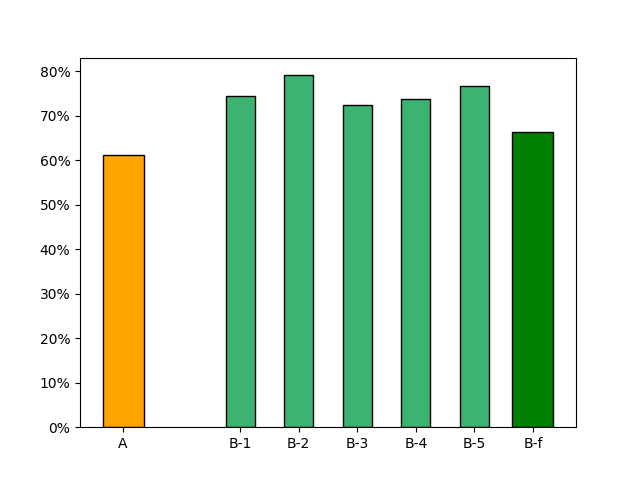}
  \caption{The results of consistency study.}
  \label{empirical2}
\end{figure}

We find that the consistency on capability points is significantly higher than the consistency of directly evaluating the overall preference, and the method of evaluating capability points first can increase the final overall consistency rate.

\section{Methodology}

\subsection{Outer Layer MoE}

\begin{figure*}[h]
  \centering
  \includegraphics[width=\linewidth]{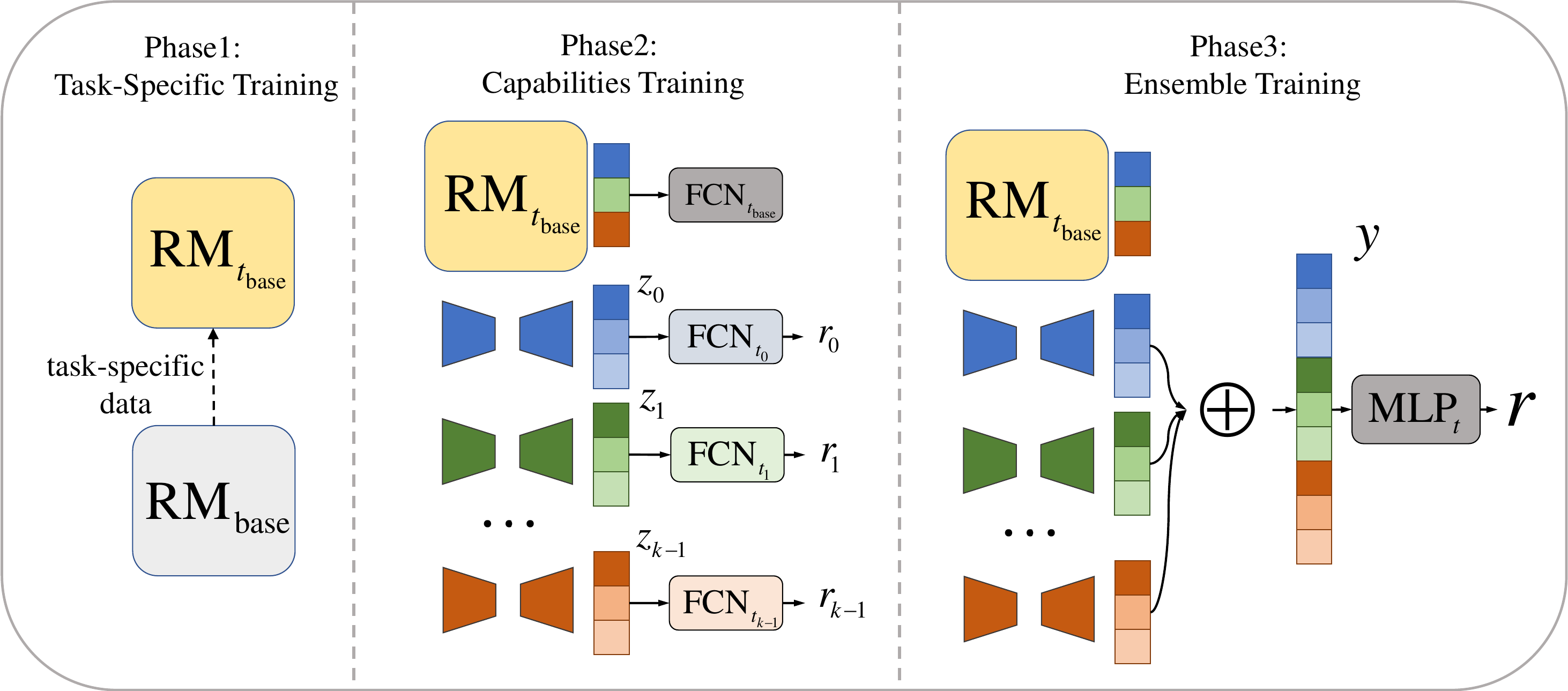}
  \caption{The training framework of each inner layer MoE. The LoRA components in the figure is only for illustration, as in actual experiments we will inject the LoRA layers into each layer of the transformers. Training details are in Section~\ref{trainingdetail}.}
  \label{framework}
\end{figure*}

Our first layer is a sparse MoE structure, and only the top-$1$ expert is activated each time. We divide the input into five categories according to tasks: text creation, roleplay, objective knowledge QA, subjective knowledge QA, and chitchat, and train an MoE for each category. When new input comes, we use a small frozen top-$1$ gating network pre-trained on category labels to act as the router. Formally, we have

\begin{equation}
	chosen = \mathop{\arg\max}_{e=0}^{4} E(t_e | x)
\end{equation}
\begin{equation}
    y = \operatorname{RM}_{chosen}(x)
\end{equation}

Here $t_0, \dots, t_4$ represent five tasks respectively, $x$ represents input, $\operatorname{RM}_0, \dots, \operatorname{RM}_4$ represent the expert RMs corresponding to each task, and $y$ represents the RM's output.
\subsection{Inner Layer MoE}

\subsubsection{Modeling}
For each task, we first obtain a base task-specific model $\operatorname{RM}_{t_\text{base}}$ from a general RM through training on task-specific preference data.
Then, we divide the tasks into distinct capability points. Capability points are equivalent to a decomposition of tasks in a low-dimensional space. Define the input space as $X$, we need to learn an expert $\operatorname{RM}_{t_i}: X \rightarrow Z_i$. In this work, we obtain $\operatorname{RM}_{t_i}$ by performing LoRA fine-tuning on $\operatorname{RM}_{t_\text{base}}$.

Let $Z = {Z_0, \dots, Z_{k-1}}$. After we have learned k experts, we use an aggregation network to aggregate the output results of k experts to produce the final output; that is, we learn $\operatorname{RM}_t: Z \rightarrow R$. This is a Markov process and $ X \rightarrow Z \rightarrow R$ construct a homogeneous Markov chain.

We employ a FCN following each expert $\operatorname{RM}{t_i}$ to serve as the value head to generate one-dimensional scores, which further maps to the range of $[0,1]$ using the sigmoid activation function. Let $W^{(base)}$ represent the initial base model $\operatorname{RM}{t_\text{base}}$, $\Delta W^{(i)}$ represent the fine-tuned LoRA network learned for capability point $i$, and $w_i$ and $b_i$ represent the FCN network associated with capability point $i$. The score $r_i$ for capability point $i$ is then expressed as follows:

\begin{equation}
    z_i = (W^{(base)} + \Delta W^{(i)})x
\end{equation}

\begin{equation}
    r_i = \sigma(w_i z_i + b_i)
\end{equation}

To obtain a single score as the final reward from multiple experts, we concatenate all low-dimensional representations of experts and use a two-layer MLP to aggregate them. Then the final reward score $r$ is:

\begin{equation}
    z = \mathop{\oplus}_{i=0}^{k-1} z_i
\end{equation}

\begin{equation}
    r = \sigma(W_1 \operatorname{PReLU}(W_0 z + B_0) + B_1)
\end{equation}

Note that our MLP does not act on the final scalar outputs, but rather on the multiple low-dimensional decomposition without being fed into FCNs, as we believe there may be underlying correlations between different capability points, which can be learned using MLP in their low-dimensional embedding space.

We use logsigmoid as the loss function, which is also the most commonly used loss function for training RMs, where $k$ represents the number of responses in a piece of data:

\begin{equation}
    \mathcal{L} = - \frac{1}{\binom{k}{2}}E_{(x,y_w,y_l) \sim D}[\log(\sigma(r(y_w)-r(y_l)))]
\end{equation}



\subsubsection{Training}
\label{trainingdetail}

Figure~\ref{framework} illustrates the training framework of each inner layer MoE. We use a pre-trained $\operatorname{RM}_{t_\text{base}}$ as the base general RM, and then perform the following three phases of training in sequence:

\begin{itemize}
  \item {\bf Phase 1: Task-Specific Training.}
    Use $60\%$ of the task-specific preference data to full-parameter fine-tune on $\operatorname{RM}_{t_\text{base}}$ to get base task-specific model  $\operatorname{RM}_{t_\text{base}}$.
    \item {\bf Phase 2: Capabilities Training.} 
    Use the data with capability point labels (the method of obtaining these labels is introduced in Section~\ref{obtaining}) to train $\operatorname{RM}_t$ using LoRA fine-tuning to obtain $\operatorname{RM}_{t_0}, \dots, \operatorname{RM}_{t_{k-1}}$. Each time, a new linear head is learned from the original linear head of $\operatorname{RM}_{t_\text{base}}$. 
    \item  {\bf Phase 3: Ensemble Training.}
    Remove the original FCN on each expert RM and use an MLP to aggregate $\operatorname{RM}_{t_0}, \dots, \operatorname{RM}_{t_k}$ to obtain the final model $\operatorname{RM}_t$, and use the remaining $40\%$ task-specific preference data to train it. During this phase, $\operatorname{RM}_{t_\text{base}}$ and the LoRA layers are frozen, and only the newly added MLP layer is trained.
\end{itemize}

\subsection{Capability Point Labels Obtaining}
\label{obtaining}

Since it is costly to obtain all preference labels on each capability point, instead of manually sorting or scoring, we use the method of calling the public Ernie Bot API\footnote{The ERNIE Bot API has similar functions to ChatGPT, but it is much cheaper and can achieve the same level of proficiency in Chinese: \url{https://cloud.baidu.com/doc/WENXINWORKSHOP/s/flfmc9do2}} to obtain the comparative preference of the single capability point in each response pair, which significantly reduces the labeling cost. 

To avoid the positional bias inherent in LLMs, we employ a strategy that involves swapping positions and requiring each pair to be processed twice. We then selectively retain only those data pairs that exhibit consistency in twice-calling. This method also doubles as a data cleansing technique, as it effectively filters out pairs with minimal discrepancies, which may introduce potential noise into training data. The prompt template is shown in both Chinese (Table~\ref{prompt-chinese}) and English (Table~\ref{propmt-english}) in the Appendix.

Our approach does not require additional data since the task-specific data from training Phases 1 and 3 can be directly utilized as raw response pairs, which enables us to acquire capability point preferences based on them. In our experiments, we reutilize all the task-specific data as raw response pairs during Phase 2.

\section{Experiment Setup}

\subsection{Model}


We use Qwen-1.8B-Chat\footnote{\url{https://huggingface.co/Qwen/Qwen-1_8B-Chat}} as both the policy model and the base reward model, which is an open source Chinese-English bilingual Transformer-based large language model proposed by Alibaba Cloud. To make a fair comparison, we use the same model as the base model of our baseline ensemble methods.

\subsection{Baseline}
\label{baseline}
We use a single RM baseline and collect three state-of-the-art ensembling methods for reward models from a range of papers
\citep{coste2023reward, eisenstein2023helping, zhai2023uncertainty}. All baseline methods and our model are trained and evaluated with the same dataset. 
\begin{itemize}
    \item {\bf Single RM} 
    
    We use the training of a single reward model as the most basic benchmark.
    \item {\bf Mean Optimization}
    
    Mean optimization simply takes the mean of the outputs of the different ensemble members:
    \begin{equation}
        R_{\mu}(x) = \frac{1}{k}\sum_{i=0}^{k-1} {r_i(x)}
     \end{equation}

    \item {\bf Worst-Case Optimization}

    Worst-case optimization (WCO) creates a conservative estimate by choosing the lowest reward from the ensemble at every step: 
    
    \begin{equation}
        R_{\text{WCO}}(x) = \min_{i=0}^{k-1} {r_i(x)}
     \end{equation}
    
    \item {\bf Uncertainty-Weighted Optimization}
    
        Uncertainty-weighted optimization (UWO) calculates reward by combining the average reward across all models in an ensemble with the intra-ensemble variance, weighted by a coefficient $\lambda$. Mathematically, this objective is given by:
        \begin{equation}
        \begin{split}
            R_{\text{UWO}}(x) = \\ \underbrace{R_{\mu}(x)}_{\text{mean}}
            &- \lambda\underbrace{\frac{1}{k}\sum_{i=0}^{k-1}(r_i(x) - R_{\mu}(x))^2}_{\text{variance}}
        \end{split}
        \end{equation}
    
\end{itemize}

\subsection{Dataset}


Our prompt dataset are very diverse and can be mainly classified into five categories: roleplay, chitchat, subjective knowledge QA, objective knowledge QA, and text creation, with some others (including logical reasoning, mathematical calculations, code understanding and generation, translation, etc). Our dataset is over $98\%$ Chinese. We ensure that the training and test sets contain no intersection and heuristically eliminate duplicate prompts. We also filter prompts containing personally identifiable information (PII).
 
Our data has a turn range of $1$ to $27$, with an average of $3.72$ (a turn contains a user query and an LLM response). Each sample's final query has multiple responses generated through either automated or manual processes. These responses are then assigned preference rankings through manual labeling. A detailed statistic of our data is shown in Table~\ref{categories}.

\begin{table}[hbtp]
  \centering
  \caption{The statistics of our dataset.}
  \label{categories}
  \begin{tabular}{c|c|c}\toprule 
  Categories & Count & Avg. Turn\\\midrule 
   text creation & $6,498$ & $2.96$ \\
   roleplay & $4,150$ & $5.32$ \\
   objective knowledge QA & $3,595$ & $3.77$ \\
   subjective knowledge QA & $3,413$ & $3.50$\\
   chitchat & $3,315$ & $3.45$\\
   others & $709$ & $3.40$ \\\midrule 
   all & $21,680$ & $3.72$ \\\bottomrule
  \end{tabular}
\end{table}

We employ a rigorous annotation process with clear documentation guides to direct our annotators (we have shown an example in our GitHub repository). Each piece of data is evaluated by two annotators to ensure quality, and the final ranking is established through discussion to reach a consensus. We have observed that the consistency rate between each pair of annotators reached $74\%$ on average.

\section{Result}

\subsection{Training Phases}

\begin{figure*}[htbp]
  \centering
  \caption{The progress of the model at different training stages. The horizontal axis of each image represents the number of training steps, and the vertical axis represents the accuracy of sorting pairs of responses on the training and testing set. Figure~\ref{fig:sub1} shows the results of the training Phase 1. Figures from~\ref{fig:sub2.1} to~\ref{fig:sub2.6} show the results of the training Phase 2. Figure~\ref{fig:sub3} (top-right) shows the results of the training Phase 3.}
  \label{phases}
  \begin{subfigure}[b]{0.44\textwidth}
    \includegraphics[width=\textwidth]{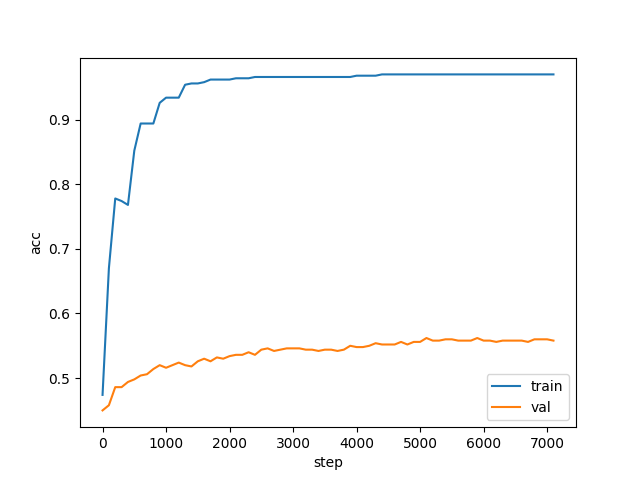}
    \caption{Phase 1}
    \label{fig:sub1}
  \end{subfigure}
  \hfill 
  \addtocounter{subfigure}{6}
  \begin{subfigure}[b]{0.44\textwidth}
    \includegraphics[width=\textwidth]{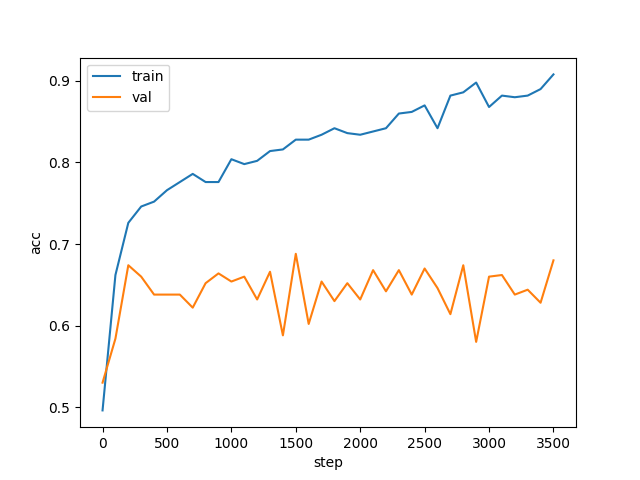}
    \caption{Phase 3}
    \label{fig:sub3}
  \end{subfigure}
  \addtocounter{subfigure}{-7}
  \begin{subfigure}[b]{0.32\textwidth}
    \includegraphics[width=\textwidth]{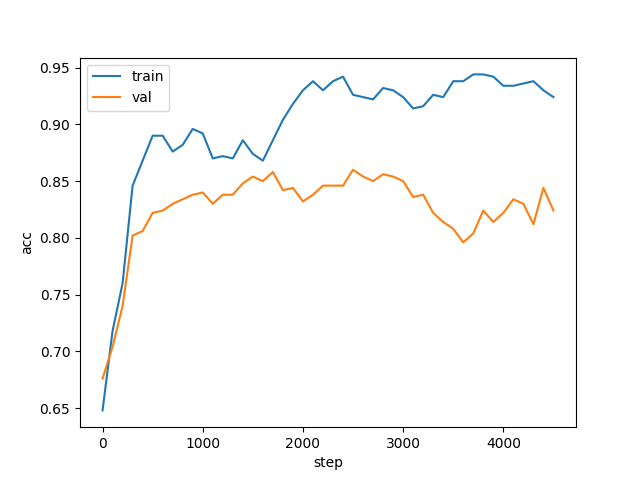}
    \caption{personality and emotional investment}
    \label{fig:sub2.1}
  \end{subfigure}
  \begin{subfigure}[b]{0.32\textwidth}
    \includegraphics[width=\textwidth]{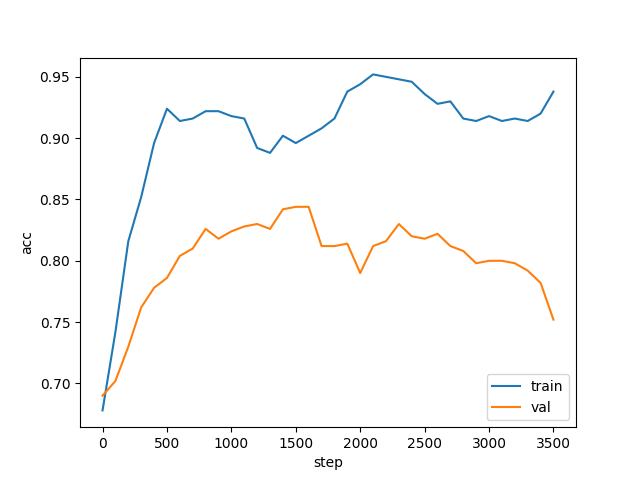}
    \caption{conversational sense}
    \label{fig:sub2.2}
  \end{subfigure}
  \begin{subfigure}[b]{0.32\textwidth}
    \includegraphics[width=\textwidth]{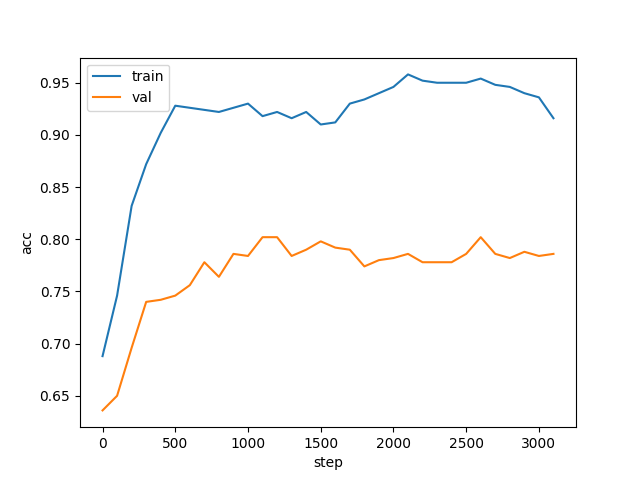}
    \caption{empathy ability}
    \label{fig:sub2.3}
  \end{subfigure}
  \begin{subfigure}[b]{0.32\textwidth}
    \includegraphics[width=\textwidth]{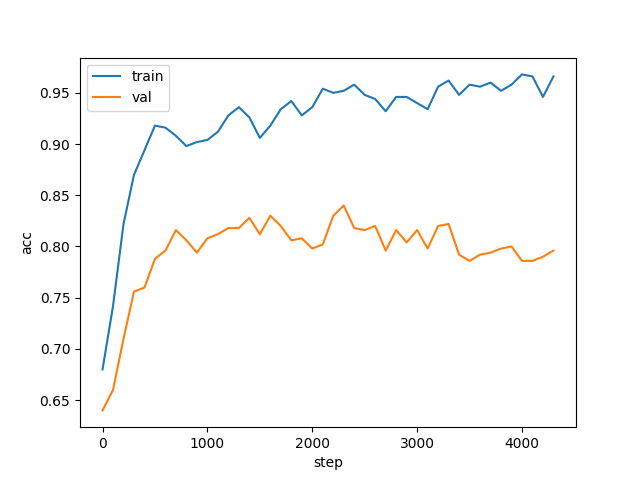}
    \caption{manifestation of relationship traits}
    \label{fig:sub2.4}
  \end{subfigure}
  \begin{subfigure}[b]{0.32\textwidth}
    \includegraphics[width=\textwidth]{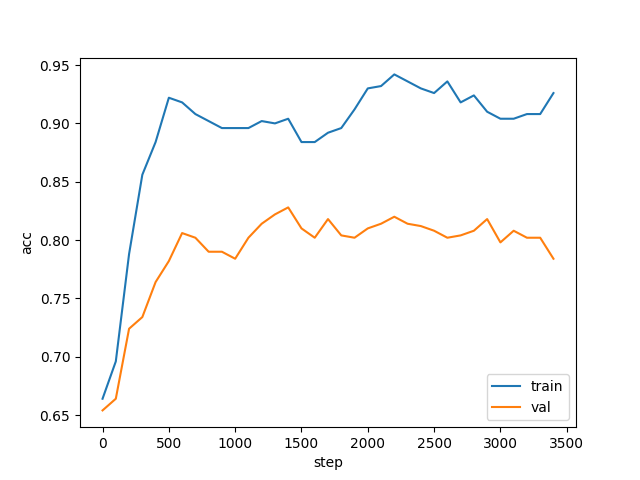}
    \caption{personalized characteristic expression}
    \label{fig:sub2.5}
  \end{subfigure}
  \begin{subfigure}[b]{0.32\textwidth}
    \includegraphics[width=\textwidth]{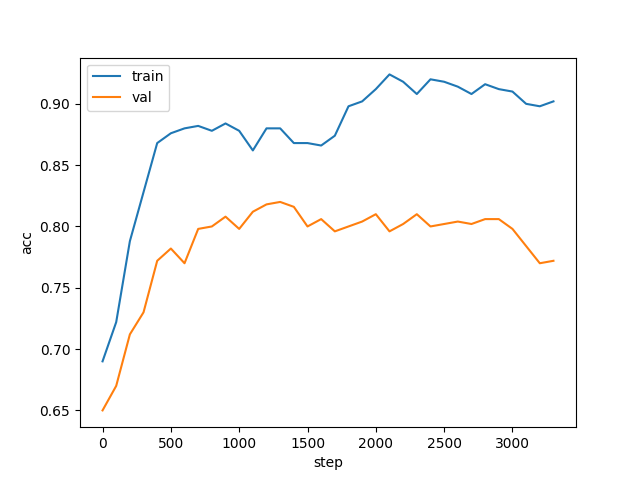}
    \caption{content richness}
    \label{fig:sub2.6}
  \end{subfigure}

\end{figure*}

\begin{table*}[hbtp]
  \centering
  \begin{tabular}{c|cccccc|c}\toprule 
  Categories & text creation & roleplay & okqa & skqa & chitchat & others & overall \\\midrule 
  Zero-shot GPT-4 & $58.9$ & $56.1$ & $64.7$ & $63.5$ & $54.8$  & $60.1$ & $59.5$ \\
  One-shot GPT-4 & $61.4$ & $59.9$ & $66.5$ & $66.2$ & $58.4$ & $\underline{61.7}$ & $62.3$ \\\midrule
   Single & $58.1$ & $57.6$ & $60.4$ & $62.0$ & $54.3$ & $54.4$ & $58.2$ \\
   Mean & $63.3$ & $62.0$ & $65.2$ & $66.3$ & $58.5$ & $56.6$ & $62.4$ \\
   WCO(min) & $63.6$ & $62.3$ & $65.5$ & $66.1$ & $59.1$ & $56.7$ & $62.6$ \\
   UWO(mean minus std) & $63.4$ & $61.8$ & $65.4$ & $66.5$ & $59.0$ & $56.6$ & $62.6$ \\\midrule
   DMoERM-w/o-Outer(\textbf{\textit{ours}}) & $\underline{68.3}$ & $\underline{66.5}$ & $\underline{70.1}$ & $\underline{71.3}$ & $\underline{63.2}$ & $61.3$ & $\underline{67.0}$ \\
   DMoERM(\textbf{\textit{ours}}) & $\mathbf{71.9}$ & $\mathbf{68.2}$ & $\mathbf{73.8}$ & $\mathbf{74.2}$ & $\mathbf{66.4}$ & $\mathbf{63.0}$ & $\mathbf{70.7}$ \\\bottomrule
  \end{tabular}
  \caption{\label{consistency} The consistency with human preferences. Note that for the methods except GPT-4, the overall results is not simply adding each up but train with all data and test. The best performance is in bold and the second best is underlined.}
\end{table*}

\begin{figure*}[htbp]
  \centering
  \caption{The optimization results for BoN and PPO for the roleplay task. The x-axes have a square-root scale, and the KL divergence scale differs between BoN and PPO due to differences in the algorithm and the KL calculation. All RMs will be normalized to have a zero mean after training.} 
  \label{optimization}
  \begin{subfigure}[b]{0.49\textwidth}
    \includegraphics[width=\textwidth]{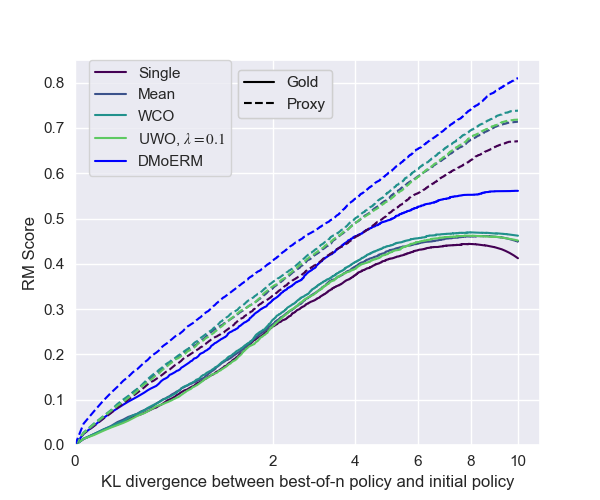}
    \caption{BoN}
    \label{fig:BoN}
  \end{subfigure}
  \begin{subfigure}[b]{0.49\textwidth}
    \includegraphics[width=\textwidth]{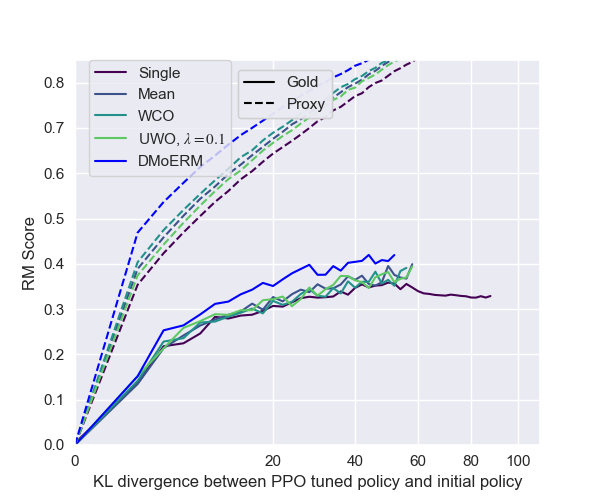}
    \caption{PPO}
    \label{fig:PPO}
  \end{subfigure}
\end{figure*}

In this section, we use roleplay as an example to demonstrate the progress of the model at different training phases. By recording the accuracy of the reward model's ranking on the pairs of responses in training and testing sets, we can intuitively display the training results. The results are shown in Figure~\ref{phases}. We find:

\begin{itemize}
    \item \textbf{In Phase 1} (Figure~\ref{fig:sub1}): the accuracy of the training set improves rapidly, while the improvement of the test set dataset is slow, eventually stabilizing at $56\%$.
    \item \textbf{In Phase 2} (Figure~\ref{fig:sub2.1} to \ref{fig:sub2.6}): we divide the roleplay task into six capability points, namely 1) personality and emotional investment, 2) conversational sense, 3) empathy ability, 4) manifestation of relationship traits, 5) personalized characteristic expression, and 6) content richness. We find that depending on different single capability points, an accuracy of $80$-$86\%$ can be achieved on the test set. Note that the training and testing set labels here are only one capability point and not an overall preference.
    \item \textbf{In Phase 3} (Figure~\ref{fig:sub3}): the improvement in the test set was rapid, reaching a peak of $68\%$. Compared with the $56\%$ accuracy in Phase 1, it can be proven that our method of training multiple experts based on different capability points and aggregating those experts can significantly improve the model's performance.
\end{itemize}

\subsection{Preference Consistency Evaluation}

Since RM is essentially an imperfect proxy for human preferences, testing the consistency rate of trained RM using human-labeled preference data is a direct and effective evaluation method. Given a pair of preference data, we use the trained RMs to assign scores to each response and record the consistency between the sorting of scores and the sorting of manual labels. Higher consistency rates (or accuracy) mean better performance of RMs as the proxies for human preferences. Compared with the methods introduced in Section~\ref{baseline}, we add the GPT-4 generative evaluation benchmarks for a more comprehensive evaluation. Their implementation details are introduced in Appendix~\ref{gpt4}.

We present the results in Table~\ref{consistency}. There are two noteworthy findings: firstly, our DMoERM achieves the best results in all categories and overall experiments, with a $6$ to $8$ percentage point improvement compared to other methods. This indicates that our training method can better learn human preferences without increasing the amount of training data or model parameters. Secondly, the DMoERM-w/o-Outer model removes the outer layer and instead uses the same capability point partition to train on various categories. It achieves the second effect and still has a significant improvement compared to other methods, making it a viable alternative when memory or task-specific data is limited.

\subsection{Optimization Evaluation}

In the optimization evaluation, we use BoN and PPO as optimization strategies to optimize the same policy model. Since we have verified in section~\ref{sec:DiffCat} that the outer MoE will improve performance by dividing different categories to train different models, for the sake of fair comparison, we restrict the task within roleplay to comparing the effects of the corresponding inner MoE with other aggregation methods. Under the same degree of optimization measured by KL divergence, we use another pre-trained reward model\footnote{\url{https://huggingface.co/IDEA-CCNL/Ziya-LLaMA-7B-Reward}} fine-tuned on our task-specific data with a larger number of parameters as the referee to judge the outputs of results, which we called the \textit{gold} reward model. Model strategies that score higher under the gold reward model are considered better. We evaluate BoN for a maximum of $n_{max} = 60,000 $ samples, which roughly equals $10$ nats of KL. The reported results represent the mean outcomes derived from a set of $500$ distinct prompts. For PPO, we train for $3,000$ PPO steps and compute the average results derived from three distinct random seeds. We give further details on implementation and other hyperparameters in Appendix~\ref{optimizationmethod}.

We showcase the results in Figure~\ref{optimization}. Both BoN and PPO optimization results demonstrate that our model consistently outperforms alternative integration approaches when policy models are optimized to the same degree. Moreover, after $n=8,000$ (KL $\approx 8$ nats) in the BoN optimization experiment, our model maintaine stability without signs of overoptimization, unlike other ensemble methods that exhibit varying degrees of overoptimization. These findings suggest that our model outperforms baselines in optimizing LLMs and is capable of addressing the issue of overoptimization.

\section{Conclusion}

In this work, we propose DMoERM to enhance RMs' performance. The outer layer MoE divides inputs into different tasks to avoid multi-task disturbance, while the inner layer MoE reduces the impact of data noise by learning LoRA experts on different capability points. Preference consistency experiments demonstrate our model is more representative of human preferences. Optimization evaluations indicate our model is more effective in optimizing LLMs and can mitigate the overoptimization problem.


\section*{Limitations}

Although we attempt to reduce annotation costs by calling public LLM API instead of manual labeling, it is still costly as using the ERNIE Bot API will cost approximately \$$3,000$ in total, and using the ChatGPT API will cost about ten times more. However, if the annotation standards for human are predetermined in advance, it will not significantly bring more annotation costs. In our annotation process, we found that due to the annotators spending most of their time understanding the queries and various responses, the proposed annotation method (first annotates the preferences of pre-determined capability points and then annotates the overall preferences) only reduces the annotator's speed by about $10\%$, but can increase the annotation consistency by $5$ percentage points.

Another problem is the training time, with about $80$ NVIDIA A100 GPU hours to train one inner MoE, which is about eight times longer than training a traditional single RM with the same amount of parameters. While many works focus on exploring efficient training methods, we reserve this problem as a future work.




\bibliography{anthology,custom}
\bibliographystyle{acl_natbib}


\appendix

\section{Additional Related Work}
\label{sec:appendix}

\subsection{Reinforcement Learning with Human Feedback}

Reinforcement Learning with Human Feedback (RLHF) is a foundational method for fine-tuning language models to align with human preferences. RLHF has been applied to a variety of tasks, including text summarization~\citep{stiennon2020learning} and improving the helpfulness and harmlessness of language models~\citep{bai2022training}. In particular, InstructGPT~\citep{2022Training} employs a three-step RLHF process that includes a supervised learning technique and the PPO algorithm~\citep{schulman2017proximal}, which has proven to be effective for ChatGPT. Despite its success, RLHF encounters several challenges, such as low sample efficiency~\citep{DBLP:conf/iclr/SnellKSYL23, DBLP:journals/corr/abs-2308-08998} and overoptimization~\citep{gao2023scaling}. Since we require no additional data to improve performance and mitigate overoptimization, our method works in both two aspects.

RLHF heavily depends on reward modeling as a proxy for human preferences. Recent research has attempted to bypass the reward modeling step~\citep{DBLP:journals/corr/abs-2304-05302-change, DBLP:journals/corr/abs-2306-17492}. Specifically, Direct Policy Optimization (DPO) aims to refine policies by classifying human preference data without reward models. Although this method is simpler to implement and offers training stability, more recent studies reveal several advantages of using reward models. Investigations into the robustness of reward-model-based strategies suggest they are more resistant to overfitting due to the limitations of KL regularization~\citep{DBLP:journals/corr/abs-2310-12036}. Moreover, in comparison to DPO, reward-model-based RLHF shows great advantages on out-of-preference samples~\citep{DBLP:journals/corr/abs-2310-10505}.

\subsubsection{the Overoptimization Problem of RMs}

As the learned reward model is only a proxy for the true reward function, optimizing it may not always result in an improvement according to true human preferences. In practice, optimizing a (fixed) learned reward model almost always leads to improvement according to this learned reward model but only improves according to the \textit{true} reward model (i.e., humans) for some initial period, after which performance often begins to regress. This phenomenon is referred to as \textit{overoptimization}.

\section{Additional Experiment Setup}

\subsection{Hyperparameters}

 Since in different training phases there are different sets of model parameters that need to be trained, we use different learning rates to better adapt to these three phases and allocate different proportions of training set for Phase 1 and Phase 3. The settings are shown in Table~\ref{hyper-phases}.

\begin{table}[h]
  \centering
  \begin{tabular}{c|c|c}\toprule 
     & learning rate & proportion \\\midrule 
   Phase 1 & $1 \times 10^{-7}$ & $60\%$ \\
   Phase 2 & $5 \times 10^{-5}$ & $-$ \\
   Phase 3 & $1 \times 10^{-6}$ & $40\%$ \\\bottomrule
  \end{tabular}
  \caption{The hyperparameters used in different training phases.}
  \label{hyper-phases}
\end{table}

We set the minibatch size to $1$ and conduct an evaluation on the validation set every $100$ training steps. If the best result on the validation set does not improve for $20$ consecutive evaluations, we implement early stopping and use the best-performing model for the next training phase or the final testing if it is already Phase 3.

The settings for the LoRA components are presented in Table~\ref{lora}, applicable to both our LoRA experts and the LoRA ensembles used in the baseline comparisons.

\begin{table}[h]
  \centering
  \begin{tabular}{l l}\toprule 
   rank  & $32$ \\\midrule 
   lora alpha & $32$ \\\midrule 
   target modules & q\_proj \& v\_proj \\\midrule 
   lora dropout & $0.05$ \\\bottomrule
  \end{tabular}
  \caption{LoRA configurations.}
  \label{lora}
\end{table}

The PPO hyperparameters and generation hyperparameters are shown in Tables~\ref{PPOhyperparameters} and~\ref{generationhyperparameters}, respectively.

\begin{table}[h]
  \centering
  \begin{tabular}{l l}\toprule 
   learning rate  & $1.41 \times 10^{-5}$ \\\midrule 
   cosine annealing scheduler & $1 \times 10^{-7}$ \\\midrule 
   PPO steps & $3,000$ \\\midrule 
   batch size & $32$ \\\midrule
   KL penalty & $0.1$ \\\midrule
   clipping range & $0.2$ \\\midrule 
   clipping value & $0.2$ \\\midrule 
   gamma  & $1.0$  \\\midrule 
   lambda & $0.95$ \\\bottomrule
  \end{tabular}
  \caption{PPO hyperparameters.}
  \label{PPOhyperparameters}
\end{table}

\begin{table}[h]
  \centering
  \begin{tabular}{l l}\toprule 
   max instruction length  & $1,024$ \\\midrule 
   max new tokens (answer length) & $1,024$ \\\midrule 
   top-p & $0.8$ \\\midrule 
   temperature & $1.0$ \\\bottomrule
  \end{tabular}
  \caption{Generation hyperparameters.}
  \label{generationhyperparameters}
\end{table}

All experiments are run on a single machine with eight NVIDIA A100 80G GPUs, and we use Adam optimizer for the optimization process.

\subsection{Ensemble Creation for Baseline Methods}

To create an ensemble for the Mean, WCO, and UWO baselines, under~\citet{coste2023reward, eisenstein2023helping} guidance, we train a fixed number of proxy reward models using identical data and hyperparameters. Each model, however, is initialized with different random seeds. This leads to variations in the random initialization of the scalar reward head that is added on top of the pre-trained language model, as well as in the data shuffling order. Table~\ref{lora} presents the LoRA parameters for these models. We have trained an ensemble of five reward models, aligning with the configurations used in previous works. This number is also comparable to the number of LoRA experts in our model, which ranges from five to six, depending on the numbers of capability points allocated under a category.

\subsection{Optimization Method}
\label{optimizationmethod}

\subsubsection{Best-of-$n$ Sampling}
        Best-of-$n$ (BoN) sampling, also called rejection sampling, is a simple inference-time optimization method~\citep{2021WebGPT, 2022Training}. For a given prompt, $n$ responses are generated from the policy model, and the answer with the highest proxy reward model score is returned. To evaluate the degree of optimization, the KL divergence~\footnote{A recent work~\citep{DBLP:journals/corr/abs-2401-01879} claims to have proved that this boundary is not so accurate and provided a tighter boundary defined by the binary entropy function. At present, they have not been widely recognized, and even if they are correct, our experimental conclusions are still valid in terms of trends.} is defined analytically as a function of $n$:

        \begin{equation}
            \operatorname{KL}_\text{bon} = \log{n} - \frac{n - 1}{n}
        \end{equation}

        In our experiments, we evaluate BoN for a maximum of $n_{max} = 60,000$ samples, which roughly equals $10$ nats of KL.
        
\subsubsection{Proximal Policy Optimization}

        Proximal Policy Optimization (PPO)~\citep{schulman2017proximal} is a policy-gradient-based online reinforcement learning method that maximizes a given reward function by repeatedly performing small incremental updates to the policy. PPO is the standard algorithm used in fine-tuning language models based on human feedback~\citep{2022Training, bai2022training, stiennon2020learning, zheng2023secrets}. When using PPO to fine-tune a language model, a KL penalty term is added during the reward calculation to regularize the policy by preventing it from deviating far from the initial policy:

        \begin{equation}
            R^{\text{PPO}}(q, r) = R(q, r) - \beta \operatorname{KL}_{\text{PPO}}
        \end{equation}
        where $\pi^{\text{PPO}}$ is the policy being optimized and $\pi^{\text{init}}$ is the initial (pre-trained) language model.
        
The naive way to calculate KL divergence between the PPO-optimized policy $\pi^{\text{PPO}}$ and the initial model is as follows:

\begin{equation}
    \operatorname{KL}_{\text{PPO}}(\pi^{\text{PPO}},\pi^{\text{init}}) = E_{(q,r)\sim \pi^{\text{PPO}}} [\log\frac{\pi^{\text{PPO}}(r|q)}{\pi^{\text{init}}(r|q)}]
\end{equation}

However, this estimator suffers from high variance and may yield negative values. Consequently, we employ the following estimator~\cite{coste2023reward}:

\begin{equation}
\begin{split} 
    \operatorname{KL}_{\text{PPO}}(\pi^{\text{PPO}},&\pi^{\text{init}}) = \\ 
    &E_{(q,r)\sim \pi^{\text{PPO}}} [\frac{1}{2} (\log\frac{\pi^{\text{PPO}}(r|q)}{\pi^{\text{init}}(r|q)})^2]
\end{split}
\end{equation}

We train a total of $3,000$ PPO steps, and the PPO parameters are shown in Table~\ref{PPOhyperparameters}.

\subsubsection{Generative Baseline}
\label{gpt4}


Apart from RM-based methods, if only the preference order of response pairs needs to be given, a simple way is to call the public language model API through zero-shot or few-shot. It can also be combined with BoN or DPO to optimize LLM. For a more comprehensive evaluation, in preference consistency experiments, we have added the GPT-4 baseline.

\begin{itemize}
    \item {\bf GPT-4}~\citep{achiam2023gpt} We use the most advanced gpt-4-1106-preview \footnote{\url{https://platform.openai.com/docs/models/gpt-4-and-gpt-4-turbo} We called the API in January 2024.} as the evaluator to evaluate a pair of preference data. Each time, we swap the positions of the responses and make two requests, and we will re-request until we get identical results from two requests as the final results.
\end{itemize}

We try to find well-prompted instruction and use both zero-shot and one-shot to evaluate and do not use few-shot for comparison, for we find it will make the context much longer and the result is not better than one-shot. There are also some other generative evaluation methods. Since they have reported in their papers that their performance is just similar with~\cite{DBLP:journals/corr/abs-2311-18702} GPT-4 or even worse than~\cite{wang2023pandalm} GPT-4, we just use GPT-4 to represent these methods.

\section{Discussion on Model Interpretability}

The single traditional RM only outputs the final reward score and we have no clues about why a response gets its score, which makes it have poor interpretability. However, our DMoERM first learns latent embedding at different capability points under specific tasks, and the final score is learned from and based on an ensemble of embeddings of different capability points. By using the FCN trained in Phase 2 to act on the corresponding hidden embedding, we can know the score on each capability point of a response. In this way, we can identify which aspects of a good response are effective and which aspects of a poor response are not. Similarly, if the overall consistency of the model is not satisfactory, we can also identify the problem and prescribe targeted solutions by analyzing the performance of each single capability point.

The method that our DMoERM checks the task of inputs, works on different capability points, and aggregates latent embeddings of each expert to obtain the final reward is just like the \textit{Chain-of-Thought} of the reward models, which can enhance the reasoning (give the reward under the context) ability of reward models as well as the interpretability of the reward models.

\section{Additional Experimental Result}

\subsection{Larger Model Performance}

We further apply the preference consistency evaluation to the Qwen7B-Chat and Qwen14B-Chat models. Except for adjusting the learning rate to adapt to different parameter sizes ([$5e-8, 2.5e-5, 5e-7$] for the 7B model and [$3e-8, 1.5e-5, 3e-7$] for the 14B model, during the three training phases), the other parameter settings and dataset partitioning are consistent with the paper, and the specific experimental results are shown in the Table~\ref{consistency-all}.

We find that our DMoERM achieved the best results among all categories and all model sizes (in bold), while DMoERM-7B achieved the second best results (is underlined). Meanwhile, in the comparison between our DMoERM in different parameter sizes, the final results for different categories improve by about $1.9\%$ for 7B and $3.6\%$ for 14B against the smallest 1.8B model, indicating that our model structure can benefit from scaling up the size of parameters. The increase in consistency rate is not too significant when scaling up due to the presence of certain noise in the labeled data (with observed $74\%$ consistency among human annotators), which leads to it being challenging to further improve when consistency rates near the $70\%$ mark. Meanwhile, due to the amount of training data is fixed, the larger models may not have been sufficiently trained, which may also be a factor contributing to the less significant improvement.

\subsection{Human Evaluation}

\begin{figure}[htbp]
  \centering
  \caption{The winning rates of DMoERM against other ensemble methods in human evaluation.}
  \label{human-eval}
  \begin{subfigure}[b]{0.45\textwidth}
    \includegraphics[width=\textwidth]{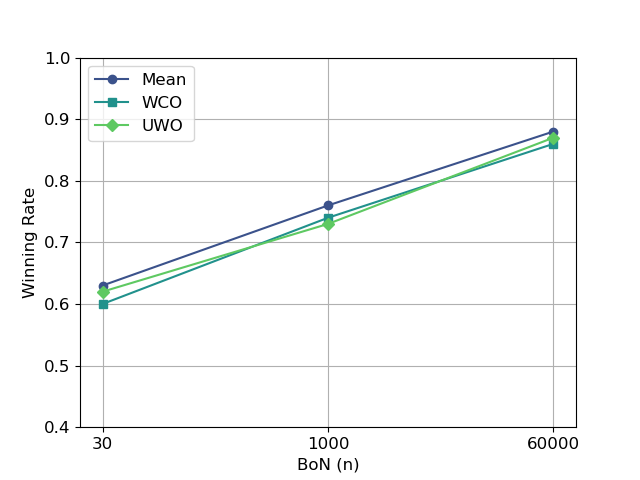}
  \end{subfigure}
  \begin{subfigure}[b]{0.45\textwidth}
    \includegraphics[width=\textwidth]{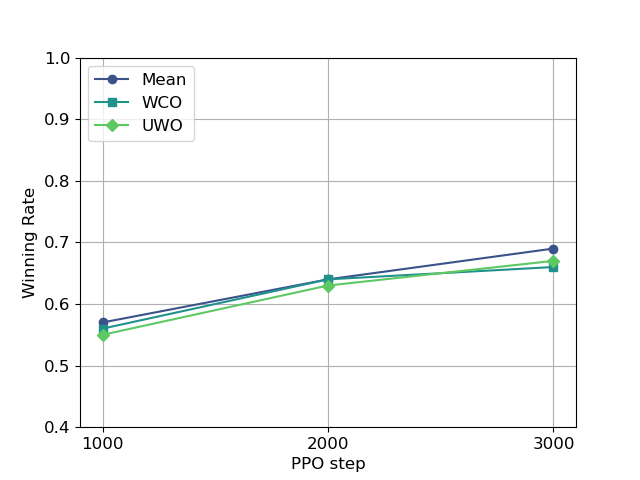}
  \end{subfigure}
\end{figure}

In the process of BoN and PPO optimization, we both set three checkpoints ($n = 30, 1000, 6000$ for BoN and $1000, 2000, 3000$ PPO steps for PPO) and use 100 prompts at each checkpoint for human evaluation. Specifically, for the different outputs obtained from these prompts under optimizations from different RMs, we require annotators to sort them according to their preferences, and we record the winning rates of our model against other ensemble methods in Figure~\ref{human-eval}.

We have two findings: Firstly, compared to BoN, the comparison improvement under PPO optimization is slower. Secondly, in both optimization experiments, the winning rates of our model increase steadily as optimization progresses, ultimately reaching approximately $87\%$ for BoN and $68\%$ for PPO.

\subsection{OOD Optimization Evaluation}

\begin{figure}[htbp]
  \centering
  \caption{The winning rates of DMoERM against other ensemble methods in OOD optimization.}
  \label{OOD-result}
  \begin{subfigure}[b]{0.45\textwidth}
    \includegraphics[width=\textwidth]{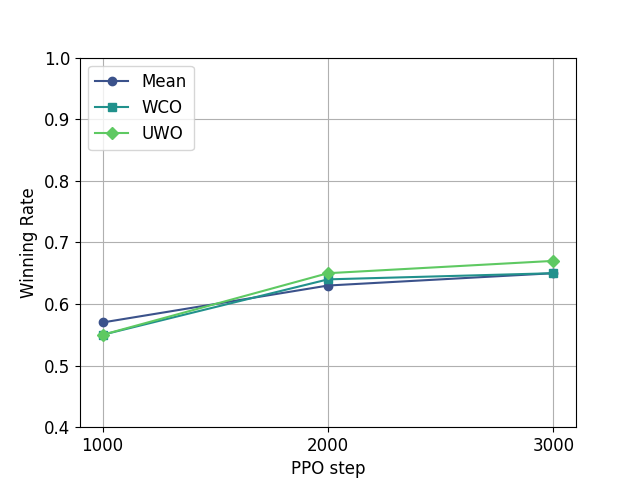}
  \end{subfigure}
  \begin{subfigure}[b]{0.45\textwidth}
    \includegraphics[width=\textwidth]{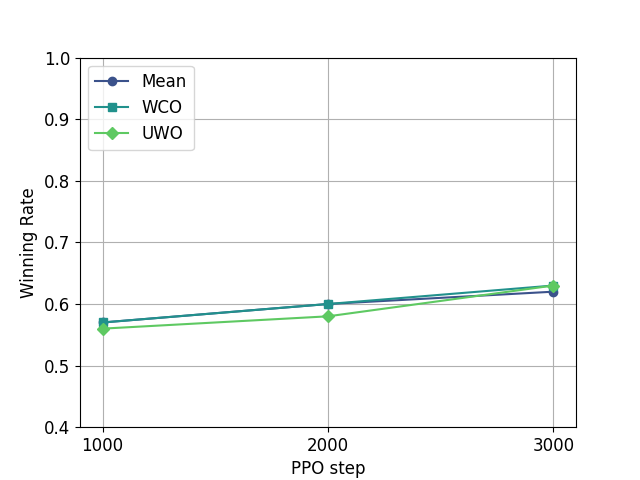}
  \end{subfigure}
\end{figure}

In contrast to several previous works~\citep{gao2023scaling, coste2023reward, eisenstein2023helping, zhai2023uncertainty} where the RM training set, optimization set, and evaluation set are typically independently and identically distributed (IID), our following experiments involve using out-of-distribution (OOD) RM training sets in PPO I and OOD evaluation sets in PPO II (since BoN is a training-free optimization method used in the inference stage, the optimization prompt set is directly the evaluation prompt set, so we only conduct OOD optimization experiments on PPO). We use AlignBench~\citep{DBLP:journals/corr/abs-2311-18743}, a publicly and popularly used alignment benchmark in Chinese, to evaluate the effectiveness of different models. We use the DMoERM-w/o-Outer for comparison and set the reward models and policy models in the same model sizes of 7B. Due to the models being larger, we compare the optimization results of different RMs using GPT-4 as the referee, which we believe has the most sufficient capability to evaluate the generation results and is naturally fair. The RM training set, the optimization set, and the evaluation set used in PPO I and PPO II are represented in Table~\ref{OOD-dataset} and the experimental results are shown in Figure~\ref{OOD-result}.

\begin{table}[ht]
\centering
\begin{tabular}{c|c|c|c}
\hline
 & \textbf{RM training} & \textbf{optimization} & \textbf{evaluation} \\ 
\hline
\textbf{I} & our dataset & AlignBench & AlignBench \\ 
\hline
\textbf{II} & our dataset & our dataset & AlignBench \\ 
\hline
\end{tabular}
\caption{The datasets used in different periods of PPO I and PPO II. Note that the datasets are randomly sampled from overall datasets and have no intersection in different periods.}
\label{OOD-dataset}
\end{table}

We have two significant findings: First, PPO I and PPO II show consistent trends, both steadily improving the winning rate against the baseline method during the optimization process. Second, In PPO II, the winning rate initially increased more rapidly (before $1,000$ steps), while the growth in PPO I was more stable and persistent in all optimization periods.

\textit{Intuitive analysis:} Due to the fact that in PPO II, the optimization set and training set are IID, the reward model can provide a clear reward signal during optimization, enabling training to quickly achieve positive optimization; conversely, In PPO I, while the optimization and evaluation sets are IID, they are OOD with the training set, making the reward signal is not such clear and strong, but advantages can gradually be reflected in the continuous optimization process.

\subsection{Annotation Quantitative Experiment}

Since our method requires more annotation costs, to provide a more comprehensive evaluation of our method, we quantified the annotation consumption and maintained consistency between our method and the baseline method in terms of annotation consumption. In this experiment, we compared DMoERM-w/o-Outer with other baseline models in all three model sizes on the entire dataset. Due to the need that our model requires each piece of data $5$ times additional annotations on intent conformity, logic, conversational sense, content richness, and readability, we only randomly sample one-sixth of the data to train our model, while other models use all the data to control the consistency of the number of annotations. The experimental results are shown in Table~\ref{annotation-cost}.

\begin{table*}[htbp]
\centering
\begin{tabular}{c|cccccc}
\hline
Model Size & Single & Mean & WCO & UWO & DMoERM-Phase1 & DMoERM-Phase3 \\
\hline
1.8B & $58.2$ & $62.4$ & $62.6$ & $62.6$ & $54.8$ & $\mathbf{64.1}$ \\
7B & $59.2$ & $63.6$ & $63.5$ & $63.6$ & $55.9$ & $\mathbf{65.3}$ \\
14B & $60.5$ & $64.9$ & $65.1$ & $65.1$ & $57.2$ & $\mathbf{66.5}$ \\
\hline
\end{tabular}
\caption{The results of labeling quantitative experiment. We record the performance of our model at the end of the training Phase 1 as DMoERM-Phase1 and at the end of three training phases as DMoERM-Phase3.}
\label{annotation-cost}
\end{table*}

We also recorded the performance of our DMoERM-w/o-Outer at the end of the training Phase 1 as DMoERM-Phase1. At this point, it is equivalent to a Single RM trained with only one-sixth of the data, so the effect is poor. But when the three training phases ended (denoted as DMoERM-Phase3), its performance improved by $9.3$ percentage points on average and exceeded all baseline models in three model sizes. 

Note that this experiment is not very fair because in our empirical experiments, we found that our annotation method only reduces the annotator's speed by about $10\%$, but can increase the annotation consistency by $5$ percentage points. In the experiment, we only used one-sixth of the data, but exceeded all the baseline methods by at least $1.4$ percentage points, indicating the compelling advantages of our method.

\subsection{Qualitative Sample}

Throughout our experiments, quantitative metrics play a pivotal role as they enable the rapid and comprehensive evaluation of various methods. In this section, we offer a concise insight into the qualitative aspect of the approaches discussed in this study. Specifically, for a given prompt, responses from the ultimate policies ($n = 60,000$ for BoN and $3,000$ steps for PPO) of each method in Figure~\ref{optimization} are provided.

The main findings are as follows. Firstly, the brevity of BoN’s response is due to the differing manner in which policy optimization occurs compared to PPO. Secondly, there are indications of failure and overoptimization in the other ensemble RMs. As shown in Table~\ref{casestudy}, for BoN, this manifests in inadequate answers. For PPO, this means bad answers that are incorrectly long and repetitive. These are obvious signs of overoptimization. Lastly, we observe that even in scenarios where other ensemble RMs struggle, our DMoERM is capable of yielding robust qualitative outcomes.

\begin{table*}[hbtp]
  \centering
  \begin{tabular}{c|c|cccccc|c}\toprule 
 Model Size & Method & text & role & okqa & skqa & chat & others & overall \\\midrule 
  & Zero-shot GPT-4 & $58.9$ & $56.1$ & $64.7$ & $63.5$ & $54.8$  & $60.1$ & $59.5$ \\
 & One-shot GPT-4 & $61.4$ & $59.9$ & $66.5$ & $66.2$ & $58.4$ & $61.7$ & $62.3$ \\  \midrule
 \multirow{6}{*}{1.8B} & Single & $58.1$ & $57.6$ & $60.4$ & $62.0$ & $54.3$ & $54.4$ & $58.2$ \\
  & Mean & $63.3$ & $62.0$ & $65.2$ & $66.3$ & $58.5$ & $56.6$ & $62.4$ \\
  & WCO(min) & $63.6$ & $62.3$ & $65.5$ & $66.1$ & $59.1$ & $56.7$ & $62.6$ \\
  & UWO(mean minus std) & $63.4$ & $61.8$ & $65.4$ & $66.5$ & $59.0$ & $56.6$ & $62.6$ \\\cline{2-9} 
  & DMoERM-w/o-Outer & $\underline{68.3}$ & $\underline{66.5}$ & $\underline{70.1}$ & $\underline{71.3}$ & $\underline{63.2}$ & $\underline{61.3}$ & $\underline{67.0}$ \\
  & DMoERM & $\mathbf{71.9}$ & $\mathbf{68.2}$ & $\mathbf{73.8}$ & $\mathbf{74.2}$ & $\mathbf{66.4}$ & $\mathbf{63.0}$ & $\mathbf{70.7}$ \\ \midrule
  
 \multirow{6}{*}{7B} & Single & $59.1$ & $58.7$ & $61.7$ & $63.2$ & $55.3$ & $55.3$ & $59.2$ \\
  & Mean & $64.7$ & $63.1$ & $66.6$ & $67.4$ & $59.4$ & $57.8$ & $63.6$ \\
  & WCO(min) & $64.4$ & $63.4$ & $66.6$ & $67.2$ & $59.3$ & $57.7$ & $63.5$ \\
  & UWO(mean minus std) & $64.4$ & $63.4$ & $66.7$ & $67.0$ & $59.3$ & $57.8$ & $63.6$ \\\cline{2-9} 
  & DMoERM-w/o-Outer & $\underline{70.0}$ & $\underline{67.9}$ & $\underline{71.6}$ & $\underline{72.9}$ & $\underline{64.5}$ & $\underline{62.4}$ & $\underline{68.4}$ \\
  & DMoERM & $\mathbf{73.2}$ & $\mathbf{69.6}$ & $\mathbf{75.2}$ & $\mathbf{75.5}$ & $\mathbf{67.6}$ & $\mathbf{64.3}$ & $\mathbf{72.1}$ \\ \midrule

 \multirow{6}{*}{14B} & Single & $60.2$ & $59.9$ & $62.7$ & $64.6$ & $56.1$ & $56.6$ & $60.5$ \\
  & Mean & $66.0$ & $64.7$ & $68.1$ & $68.7$ & $60.6$ & $58.7$ & $64.9$ \\
  & WCO(min) & $66.3$ & $64.6$ & $68.4$ & $68.5$ & $60.6$ & $58.8$ & $65.1$ \\
  & UWO(mean minus std) & $66.0$ & $64.7$ & $68.1$ & $68.6$ & $60.6$ & $58.8$ & $65.1$ \\\cline{2-9} 
  & DMoERM-w/o-Outer & $\underline{70.8}$ & $\underline{69.4}$ & $\underline{73.0}$ & $\underline{73.9}$ & $\underline{65.7}$ & $\underline{63.7}$ & $\underline{69.6}$ \\
  & DMoERM & $\mathbf{74.7}$ & $\mathbf{71.2}$ & $\mathbf{76.7}$ & $\mathbf{77.0}$ & $\mathbf{69.1}$ & $\mathbf{65.5}$ & $\mathbf{73.2}$ \\ \midrule
  \end{tabular}
  \caption{\label{consistency-all} The consistency of different model sizes with human preferences. In each model size, the best performance is in bold and the second best is underlined. We can observe that our DMoERM and DMoERM-w/o-Outer consistently achieve the best and second-best results at different model sizes.}
\end{table*}

\clearpage

\newcommand{\tabincell}[2]{\begin{tabular}{@{}#1@{}}#2\end{tabular}}

\begin{table*}[h] 
\begin{center} 
\caption{A set of example answers to an evaluation query.}
\label{casestudy}

\begin{tabular}{c|p{6.5cm}|p{6.5cm}} 
\hline
\rule{0pt}{12pt}
\tabincell{c}{Query} & \begin{tabular}[t]{@{}p{6.5cm}@{}}
\begin{CJK*}{UTF8}{gbsn}如何简洁地描写点心的香味从盒子里飘出\end{CJK*}
\end{tabular}
&
\begin{tabular}[t]{@{}p{6.5cm}@{}}
How to briefly describe the aroma of the dim sum floating out of the box
\end{tabular}

\\\hline
\rule{0pt}{12pt}
\tabincell{c}{Single\\(BoN)} & \begin{tabular}[t]{@{}p{6.5cm}@{}}
\begin{CJK*}{UTF8}{gbsn}当你打开点心盒时，那种香气会像风筝一样飘出来，让你的味蕾仿佛跳上了云朵，让你的心情变得愉快。\end{CJK*}
\end{tabular}
&
\begin{tabular}[t]{@{}p{6.5cm}@{}}
{\small When you open the snack box, the aroma will float out like a kite, making your taste buds seem to jump on the clouds and making you feel happy.}
\end{tabular}

\\\hline
\rule{0pt}{12pt}
\tabincell{c}{Mean\\(BoN)} & \begin{tabular}[t]{@{}p{6.5cm}@{}}
\begin{CJK*}{UTF8}{gbsn}打开点心盒，那怡人的香气就像一束光照亮了整个房间，仿佛能够看到它们在空气中跳跃，让人不禁垂涎三尺。\end{CJK*}
\end{tabular}
&
\begin{tabular}[t]{@{}p{6.5cm}@{}}
{\small Open the box of refreshments, the pleasant aroma lights up the whole room like a beam of light, as if you can see them jumping in the air, which makes people can't help salivating.}
\end{tabular}

\\\hline
\rule{0pt}{12pt}
\tabincell{c}{WCO\\(BoN)} & \begin{tabular}[t]{@{}p{6.5cm}@{}}
\begin{CJK*}{UTF8}{gbsn}一股香气从盒子里飘出，那是点心的香味，让人的口水不禁流了出来，欲望在空气中凝结。\end{CJK*}
\end{tabular}
&
\begin{tabular}[t]{@{}p{6.5cm}@{}}
{\small A fragrance floats out of the box, which is the aroma of refreshments, making people's saliva flow out, and desire condenses in the air.}
\end{tabular}

\\\hline
\rule{0pt}{12pt}
\tabincell{c}{UWO\\(BoN)} & \begin{tabular}[t]{@{}p{6.5cm}@{}}
\begin{CJK*}{UTF8}{gbsn}打开盒子的一刹那，点心的香气就如同一缕春风，绕梁三日，飘散在空气中，令人垂涎欲滴。\end{CJK*}
\end{tabular}
&
\begin{tabular}[t]{@{}p{6.5cm}@{}}
{\small The instant I opened the box, the aroma of the dessert spread in the air like a wisp of spring breeze lingering around for three days, which made people's mouths water.}
\end{tabular}

\\\hline
\rule{0pt}{12pt}
\tabincell{c}{DMoERM\\(BoN)} & \begin{tabular}[t]{@{}p{6.5cm}@{}}
\begin{CJK*}{UTF8}{gbsn}点心香气从盒中四溢，诱人味蕾，满室生香，令人陶醉。\end{CJK*}
\end{tabular}
&
\begin{tabular}[t]{@{}p{6.5cm}@{}}
{\small 
The aroma of dim sum overflows from the box, tempts the taste buds, fills the room with fragrance, and intoxicates people.}
\end{tabular}

\\\hline
\rule{0pt}{12pt}
\tabincell{c}{Single\\(PPO)} & \begin{tabular}[t]{@{}p{6.5cm}@{}}
\begin{CJK*}{UTF8}{gbsn}点心盒子里隐藏着一种难以言表的诱惑，那是甜点所独有的甜美芳香。随着盒盖的轻启，这股香味仿佛被赋予了生命，它不再受限于那狭小的空间，开始缓缓地向四周扩散。它如同一个害羞的少女，初出茅庐，带着些许的迟疑和羞涩，一点一点地探出头来。先是一丝若有若无的甜腻，让人心生疑惑，随即又增强了几分，变得明确而肯定。它似乎在空气中跳跃、旋转，与每一个嗅觉细胞亲密接触，勾起了人们内心深处对甜蜜的向往。\end{CJK*}
\end{tabular}
&
\begin{tabular}[t]{@{}p{6.5cm}@{}}
{\small 
The dim sum box hides an ineffable temptation, a unique sweet aroma that belongs exclusively to desserts. With the lightest lifting of the lid, this fragrance seems to be endowed with life, no longer confined to its narrow space and begins to slowly diffuse outwards. Like a shy maiden emerging from her shell, it tentatively and bashfully peeks out, little by little. At first, it's a mere hint of sweetness that leaves one puzzled, but it gradually strengthens and becomes clearer and more definite. It seems to dance and swirl in the air, intimately caressing every olfactory cell, evoking a deep-seated yearning for sweetness within people's hearts.}
\end{tabular}

\\\hline

\multicolumn{3}{r}{\textit{Continued on the next page}}

\end{tabular}

\end{center}
\end{table*}

\clearpage

\begin{table*}[h] 
\begin{center} 

\begin{tabular}{c|p{6.5cm}|p{6.5cm}} 
\hline
\rule{0pt}{12pt}
\tabincell{c}{Query} & \begin{tabular}[t]{@{}p{6.5cm}@{}}
\begin{CJK*}{UTF8}{gbsn}如何简洁地描写点心的香味从盒子里飘出\end{CJK*}
\end{tabular}
&
\begin{tabular}[t]{@{}p{6.5cm}@{}}
How to briefly describe the aroma of the dim sum floating out of the box
\end{tabular}

\\\hline
\rule{0pt}{12pt}
\tabincell{c}{Mean\\(PPO)} & \begin{tabular}[t]{@{}p{6.5cm}@{}}
\begin{CJK*}{UTF8}{gbsn}盒子轻轻打开，点心的香气瞬间弥漫。那是糯米的温润，豆沙的甜蜜，还有一丝丝清香的茶叶味。一阵阵，一股股，绵绵不绝，像是诗一般的甜美旋律，缭绕在空气中，点心的清香让人垂涎欲滴，心生欢喜。\end{CJK*}
\end{tabular}
&
\begin{tabular}[t]{@{}p{6.5cm}@{}}
{\small 
The box opens gently, and the aroma of the dessert instantly spreads. It is the warmth of glutinous rice, the sweetness of red bean paste, and a trace of fresh tea flavor. Wave upon wave, continuous and unending, like a poetic sweet melody lingering in the air, the fragrance of the dessert makes people salivate and feel delighted.}
\end{tabular}

\\\hline
\rule{0pt}{12pt}
\tabincell{c}{WCO\\(PPO)} & \begin{tabular}[t]{@{}p{6.5cm}@{}}
\begin{CJK*}{UTF8}{gbsn}
点心的香味从盒子里缓缓飘出，一次又一次地扑鼻而来，像是阵阵潮汐，持续不断地冲击着味蕾。每一波香气都带着诱人的酥脆和甜蜜，让人忍不住想要尝一口。这就是点心的魅力，它的香味像是魔法，一次又一次地吸引着人们。\end{CJK*}
\end{tabular}
&
\begin{tabular}[t]{@{}p{6.5cm}@{}}
{\small The aroma of the dim sum slowly wafts out of the box, hitting the nostrils over and over again, like waves of the tide, constantly impacting the taste buds. Each wave of fragrance brings tempting crunch and sweetness, making it hard to resist taking a bite. This is the charm of dim sum, its fragrance is like magic, attracting people over and over again.}
\end{tabular}

\\\hline
\rule{0pt}{12pt}
\tabincell{c}{UWO\\(PPO)} & \begin{tabular}[t]{@{}p{6.5cm}@{}}
\begin{CJK*}{UTF8}{gbsn}点心的香气，如同精灵的歌唱，娓娓地从盒子的缝隙中溢出。一次，两次，每一次都像是在空气中划出一个甜美的弧线。轻轻一闻，那香气又再次被唤醒，从盒子里流淌出来，一次又一次，像是永无止境的旋律，让人沉醉。\end{CJK*}
\end{tabular}
&
\begin{tabular}[t]{@{}p{6.5cm}@{}}
{\small 
The aroma of the dim sum, like the singing of elves, spills out of the cracks of the box in a continuous stream. Once, twice, each time it seems to draw a sweet arc in the air. With a gentle sniff, the aroma is awakened again, flowing out of the box, over and over again, like an endless melody that intoxicates the senses.}
\end{tabular}

\\\hline
\rule{0pt}{12pt}
\tabincell{c}{DMoERM\\(PPO)} & \begin{tabular}[t]{@{}p{6.5cm}@{}}
\begin{CJK*}{UTF8}{gbsn}点心的香气如轻烟，缓缓从精致的盒子中逸出，弥漫在空气中。那甜美的芬芳，仿佛有魔力般，无声地诱惑着每一个嗅觉细胞，让人不由自主地靠近，沉醉在这份难以抗拒的美味预告中。\end{CJK*}
\end{tabular}
&
\begin{tabular}[t]{@{}p{6.5cm}@{}}
{\small 
The aroma of the dim sum, like light smoke, escapes slowly from the exquisite box and pervades the air. That sweet fragrance, as if magical, silently tempts every olfactory cell, drawing people irresistibly closer, intoxicated by this irresistible preview of deliciousness.}
\end{tabular}

\\\hline

\end{tabular}

\end{center}
\end{table*}

\clearpage

\begin{figure*}[h]
  \centering
  \includegraphics[width=\linewidth]{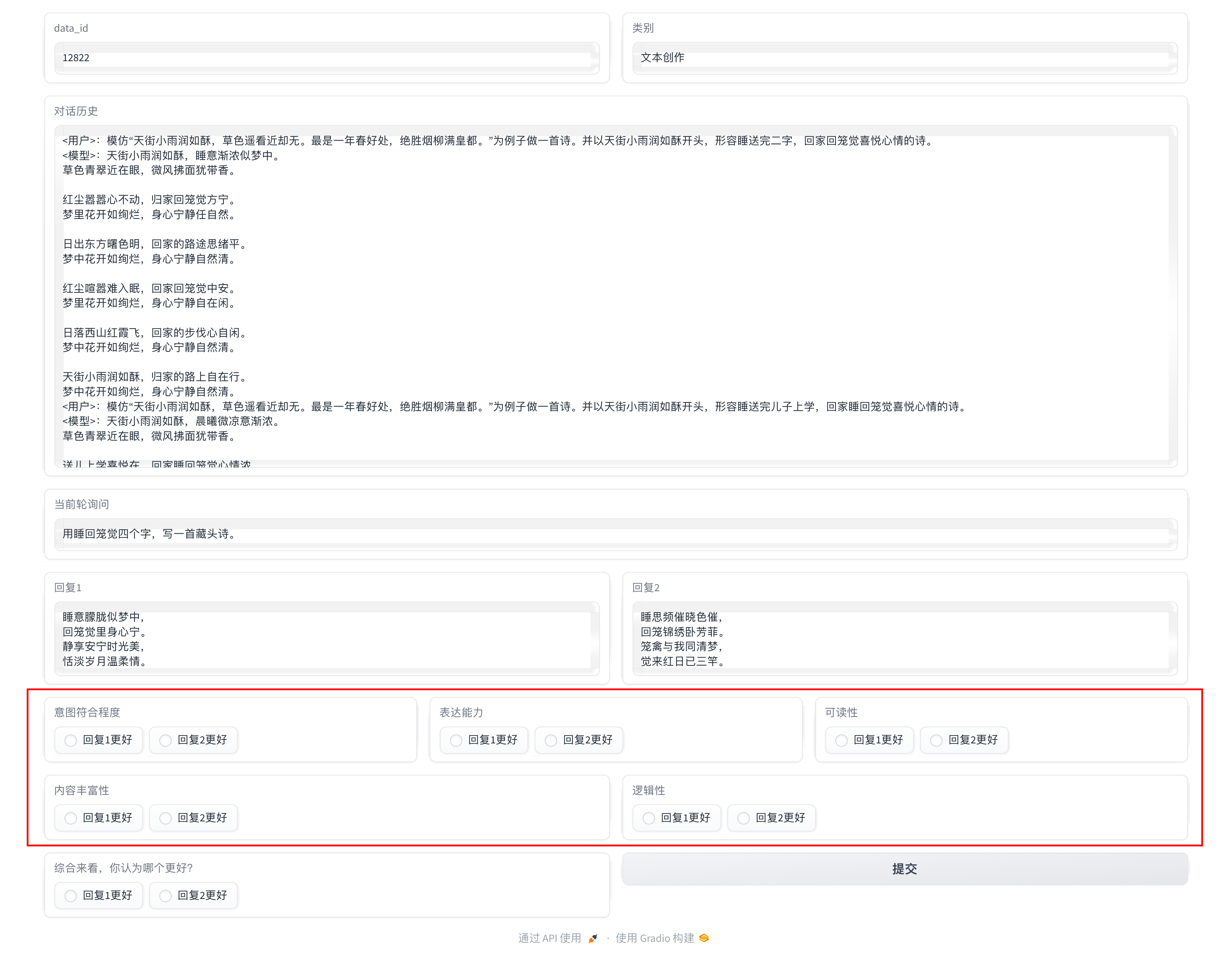}
  \caption{The screenshot of the interface for the annotation consistency experiment. The content in the red box is the preference selection for the five capability points. We find that removing them would lead to a drop of $5$ percentage points in the overall consistency of preferences.}
  \label{screenshot}
\end{figure*}

\clearpage

\renewcommand{\arraystretch}{1.5}
\begin{table*}
\centering
\begin{tabular}{c|c|c}
\hline
Category & Capability Points & Chinese \\
\hline
\multirow{6}{*}{roleplay} & personality and emotional investment & \begin{CJK*}{UTF8}{gbsn}人设和情感带入\end{CJK*} \\ \cline{2-3} 
                   & conversational sense & \begin{CJK*}{UTF8}{gbsn}对话感\end{CJK*} \\ \cline{2-3} 
                   & empathy ability & \begin{CJK*}{UTF8}{gbsn}共情能力\end{CJK*} \\ \cline{2-3} 
                   & manifestation of relationship traits & \begin{CJK*}{UTF8}{gbsn}关系特点体现\end{CJK*} \\ \cline{2-3}
                   & personalized characteristic expression & \begin{CJK*}{UTF8}{gbsn}个性化特征体现\end{CJK*} \\  \cline{2-3}
                   & content richness & \begin{CJK*}{UTF8}{gbsn}内容丰富性\end{CJK*} \\ \hline 
\multirow{5}{*}{chitchat} & conversational sense & \begin{CJK*}{UTF8}{gbsn}对话感\end{CJK*} \\ \cline{2-3}
                      & proactivity & \begin{CJK*}{UTF8}{gbsn}主动性\end{CJK*} \\ \cline{2-3}
                      & emotion expression & \begin{CJK*}{UTF8}{gbsn}情感表达\end{CJK*} \\ \cline{2-3}
                      & empathy ability & \begin{CJK*}{UTF8}{gbsn}共情能力\end{CJK*} \\ \cline{2-3}
                      & content richness & \begin{CJK*}{UTF8}{gbsn}内容丰富性\end{CJK*} \\ \hline

\multirow{5}{*}{subjective knowledge QA} & convincing ability & \begin{CJK*}{UTF8}{gbsn}说服力\end{CJK*} \\ \cline{2-3}
                          & logic & \begin{CJK*}{UTF8}{gbsn}逻辑性\end{CJK*} \\ \cline{2-3}
                          & viewpoint richness & \begin{CJK*}{UTF8}{gbsn}观点丰富度\end{CJK*} \\ \cline{2-3}
                          & breadth of knowledge & \begin{CJK*}{UTF8}{gbsn}知识面广度\end{CJK*} \\ \cline{2-3}
                          & question specific & \begin{CJK*}{UTF8}{gbsn}问题针对性\end{CJK*} \\ \hline

\multirow{6}{*}{objective knowledge QA} & correctness & \begin{CJK*}{UTF8}{gbsn}正确性\end{CJK*} \\ \cline{2-3}
                         & objectiveness & \begin{CJK*}{UTF8}{gbsn}客观程度\end{CJK*} \\ \cline{2-3}
                         & reasoning ability & \begin{CJK*}{UTF8}{gbsn}推理能力\end{CJK*} \\ \cline{2-3}
                         & logic & \begin{CJK*}{UTF8}{gbsn}逻辑性\end{CJK*} \\ \cline{2-3}
                         & depth of knowledge & \begin{CJK*}{UTF8}{gbsn}知识面深度\end{CJK*} \\ \cline{2-3}
                         & question specific & \begin{CJK*}{UTF8}{gbsn}问题针对性\end{CJK*} \\ \hline

\multirow{5}{*}{text creation} & intent conformity & \begin{CJK*}{UTF8}{gbsn}意图符合程度\end{CJK*} \\ \cline{2-3}
                      & expressiveness & \begin{CJK*}{UTF8}{gbsn}表达能力\end{CJK*} \\ \cline{2-3}
                      & readability & \begin{CJK*}{UTF8}{gbsn}可读性\end{CJK*} \\ \cline{2-3}
                      & content richness & \begin{CJK*}{UTF8}{gbsn}内容丰富性\end{CJK*} \\ \cline{2-3}
                      & logic & \begin{CJK*}{UTF8}{gbsn}逻辑性\end{CJK*} \\ \hline

\multirow{5}{*}{\begin{tabular}{@{}c@{}}overall\\(for DMoERM-w/o-Outer)\end{tabular}} & intent conformity & \begin{CJK*}{UTF8}{gbsn}意图符合程度\end{CJK*} \\ \cline{2-3}
                      & logic & \begin{CJK*}{UTF8}{gbsn}逻辑性\end{CJK*} \\ \cline{2-3}
                      & conversational sense & \begin{CJK*}{UTF8}{gbsn}对话感\end{CJK*} \\ \cline{2-3}
                      & content richness & \begin{CJK*}{UTF8}{gbsn}内容丰富性\end{CJK*} \\ \cline{2-3}
                      & readability & \begin{CJK*}{UTF8}{gbsn}可读性\end{CJK*} \\ \hline

\end{tabular}
\caption{The capability points partitions for each task in our experiments.}
\label{capabilitypartition}
\end{table*}

\clearpage

\begin{table*}[h] 
\begin{center} 
\begin{tabular}{l|p{13.9cm}} 
\hline
Template & \begin{tabular}[t]{@{}p{13.9cm}@{}}
\begin{CJK*}{UTF8}{gbsn}给定提问和两个回复，判断回复1和回复2在【\{point\}】方面哪个更好。\end{CJK*}\\
\begin{CJK*}{UTF8}{gbsn}【提问】：\{query\}\end{CJK*}\\
\begin{CJK*}{UTF8}{gbsn}【回复1】：\{response1\}\end{CJK*}\\
\begin{CJK*}{UTF8}{gbsn}【回复2】：\{response2\}\end{CJK*}\\
\end{tabular} \\
\hline
Example & \begin{tabular}[t]{@{}p{13.9cm}@{}}
\begin{CJK*}{UTF8}{gbsn}给定提问和两个回复，判断回复1和回复2在【对话感】方面哪个更好。\end{CJK*}\\
\begin{CJK*}{UTF8}{gbsn}【提问】：你觉得我要不要裸辞\end{CJK*}\\
\begin{CJK*}{UTF8}{gbsn}【回复1】：裸辞和被公司优化后离职有很大的不同，具体取决于个人情况和职业规划。以下是一些需要考虑的因素：

1. 职业规划：如果你的职业规划中包括了一份长期的工作，那么裸辞可能会影响到你未来的职业发展。在这种情况下，你需要认真思考自己的职业规划，并考虑如何在不影响自己长远发展的前提下处理眼前的问题。
2. 经济压力：离职后可能需要面对经济上的压力，例如房租、信用卡账单等。如果你的经济状况不是很好，可以先找一份短期工作，缓解一下经济压力。但是需要注意的是，短期工作可能无法满足你的职业发展需求，也不利于建立长期的职业规划。
3. 心理压力：离职后可能会带来一定的心理压力，例如失去工作、社交圈子等方面的变化等。如果你感到无法承受这种压力，可以考虑暂时休息或寻求心理咨询帮助。
4. 个人成长：在一份工作中，可以不断学习和成长，积累经验和技能，这对于未来的职业发展也是非常有帮助的。因此，即使面临经济压力，也需要尽可能地保持积极的心态，寻求更好的发展机会。

总之，裸辞和被公司优化后离职都需要慎重考虑，需要综合考虑自己的职业规划、经济压力、心理压力等多方面因素。如果决定离职，最好提前做好规划，尽可能减少离职对自己的影响。\end{CJK*}\\
\begin{CJK*}{UTF8}{gbsn}【回复2】：考虑到你的困扰，是否选择裸辞确实是一个需要深思熟虑的决定。在回答这个问题之前，我想先了解一下你目前的经济状况，是否有足够的储蓄来应对离职后可能的经济压力？另外，你的职业规划中，是否有更合适的工作机会在等待你？还有，你目前的心理状况能否承受裸辞带来的压力？

裸辞的好处是，你可以暂时摆脱目前工作环境中的压力，给自己一些休息和反思的时间。然而，它的风险在于可能会给你带来经济上的压力，并且在你寻找下一份工作的过程中，可能会出现职业空窗期，这对你未来的职业发展可能会有一些影响。

所以，我建议你在做决定之前，尽量找一些职业咨询专家进行咨询，他们可以根据你的具体情况提供更具体的建议。同时，如果你感到压力很大，也可以考虑寻求心理咨询的帮助。\end{CJK*}\\
\hline
\begin{CJK*}{UTF8}{gbsn}
在对话感方面，回复2比回复1更好。

回复1虽然详细列出了一系列关于是否应裸辞的因素让提问者考虑，但更像是一篇分析文章或建议清单，缺少对话的互动性和针对性。此外，回复1没有直接回应提问者的情感或具体状况，而是给出了一般性的建议，这在对话中可能显得有些疏远。

相比之下，回复2首先从提问者的困扰出发，提出了一些具体问题来了解提问者的状况，这种方式更贴近对话的形式，有助于建立共鸣和理解。接着，回复2也分析了裸辞的利弊，并在最后给出了具体的建议，如咨询职业咨询专家或寻求心理咨询帮助。这些建议更具体，也更具操作性，显示出回复者对提问者问题的深度理解。

所以，回复2在对话感方面表现更优秀。\end{CJK*}\\
\end{tabular} \\
\hline
\end{tabular}

\end{center}
\caption{The Chinese original version of the prompt for calling public LLM API to get the comparison on a single capability point. Note that we do not attach historical information, for it can greatly reduce the cost and is sufficient to achieve satisfactory results in our experiments. We determine the model's choice by identifying whether the number 1 or 2 appears first in its response.}
\label{prompt-chinese}
\end{table*}

\begin{table*}[h] 
\begin{center} 
\begin{tabular}{l|p{13.9cm}} 
\hline
Template & \begin{tabular}[t]{@{}p{13.9cm}@{}}
{\small Given a query and two responses, determine which of response 1 and response 2 is better in terms of [\{point\}].\par}
{\small [Query]: \{query\} \par}
{\small [Response 1]: \{response1\} \par}
{\small [Response 2]: \{response2\} \par}
\end{tabular} \\
\hline
Example & \begin{tabular}[t]{@{}p{13.9cm}@{}}
{\small Given a query and two responses, determine which of Response 1 and Response 2 is better in terms of [conversational sense].\par}
{\small [Query]: Do you think I should quit my job without having another one in hand \par}
{\small [Response 1]: Quitting a job without a backup plan and being laid off as part of a company's restructuring are quite different and depend on personal circumstances and career planning. Here are some factors to consider:

1. Career Planning: If your career planning includes a long-term job, quitting abruptly might affect your future career development. In this case, you should carefully consider your career plans and how to address current issues without jeopardizing your long-term prospects.
2. Financial Pressure: After leaving a job, you may face financial pressures such as rent or credit card bills. If you are not in a good financial situation, you might consider finding a short-term job to alleviate financial stress. However, keep in mind that short-term jobs may not meet your career development needs and may not be conducive to establishing a long-term career plan.
3. Psychological Pressure: Resigning might bring certain psychological pressures, such as the loss of a job and changes in your social circle. If you feel unable to cope with this pressure, consider taking a break or seeking psychological counseling for support.
4. Personal Growth: In a job, you can continuously learn and grow, accumulating experience and skills, which is very beneficial for future career development. Therefore, even in the face of financial pressure, it is necessary to maintain a positive attitude and seek better development opportunities.

In summary, quitting a job without a plan and being laid off due to company restructuring both require careful consideration, taking into account factors such as your career planning, financial pressure, and psychological stress. If you decide to leave, it is best to plan ahead to minimize the impact on yourself.\par}
{\small 
[Response 2]: Considering your concerns, deciding whether to resign without another job offer in hand is indeed a decision that requires careful deliberation. Before answering this question, I'd like to first understand your current financial situation. Are you equipped with enough savings to handle the potential financial pressures after resigning? Additionally, are there more suitable job opportunities waiting for you in your career plan? And, are you mentally prepared to handle the stress that may come with resigning without a backup plan?

The advantage of resigning without a backup plan is that you can temporarily escape the pressures of your current work environment, allowing yourself some time to rest and reflect. However, the risk is that it can lead to financial stress, and during the process of finding your next job, there may be a gap in your career which could impact your future professional development.

Therefore, before making a decision, I suggest that you consult with career counseling experts who can provide more specific advice based on your situation. Also, if you are feeling significant stress, consider seeking help from a mental health professional.\par}\\
\hline
{\small 
In terms of conversational sense, Response 2 is better than Response 1.

Response 1, while detailing a list of factors for the questioner to consider regarding whether to quit without a backup plan, reads more like an analytical article or a list of suggestions, lacking interactive and targeted dialogue. Moreover, Response 1 did not directly address the questioner's emotions or specific circumstances; instead, it offered general advice, which may seem somewhat distant in a conversation. 

By contrast, Response 2 begins by addressing the questioner's concerns and poses specific questions to understand their situation, which is closer to the form of a dialogue and helps to establish empathy and understanding. Then, Response 2 also analyzes the pros and cons of quitting without a plan and concludes with concrete suggestions, such as consulting career counseling experts or seeking mental health counseling. These suggestions are more specific and actionable, showing a deeper understanding of the questioner's issue.

Therefore, Response 2 performs better in terms of conversational sense.\par}\\
\end{tabular} \\
\hline
\end{tabular}

\end{center}
\caption{The English translated version of the prompt for calling public LLM API to get the comparison on a single capability point. Note that we do not attach historical information, for it can greatly reduce the cost and is sufficient to achieve satisfactory results in our experiments. We determine the model's choice by identifying whether the number 1 or 2 appears first in its response.}
\label{propmt-english}
\end{table*}

\end{document}